\documentclass[preprint,12pt,authoryear]{elsarticle}

\usepackage{amssymb}
\usepackage[unicode]{hyperref}

\hypersetup{
  pdftitle   = {From 2D to 3D, Deep Learning-based Shape Reconstruction in Magnetic Resonance Imaging: A Review},
  pdfauthor  = {Emma McMillian; Abhirup Banerjee; Alfonso Bueno-Orovio},
  pdfkeywords= {magnetic resonance imaging, deep learning, 3D reconstruction}
}

\usepackage{amsmath}
\usepackage{longtable}
\usepackage{xurl}

\journal{Arxiv}

\begin{document}

\begin{frontmatter}



\title{From 2D to 3D, Deep Learning-based Shape Reconstruction in Magnetic Resonance Imaging: A Review}

\author[1]{Emma McMillian}
\author[2]{Abhirup Banerjee}
\author[1]{Alfonso Bueno-Orovio\corref{cor1}}
\ead{alfonso.bueno@cs.ox.ac.uk} 

\cortext[cor1]{Corresponding author.}

\affiliation[1]{organization={University of Oxford},
  addressline={Department of Computer Science, Parks Road},
  city={Oxford},
  postcode={OX1 3QG},
  country={United Kingdom}
}
\affiliation[2]{organization={University of Oxford},
  addressline={Department of Engineering Science, Institute of Biomedical Engineering},
  city={Oxford},
  postcode={OX3 7DQ},
  country={United Kingdom}
}

\begin{abstract}
Deep learning-based 3-dimensional (3D) shape reconstruction from 2-dimensional (2D) magnetic resonance imaging (MRI) has become increasingly important in medical disease diagnosis, treatment planning, and computational modeling. This review surveys the methodological landscape of 3D MRI reconstruction, focusing on 4 primary approaches: point cloud, mesh-based, shape-aware, and volumetric models. For each category, we analyze the current state-of-the-art techniques, their methodological foundation, limitations, and applications across anatomical structures. We provide an extensive overview ranging from cardiac to neurological to lung imaging. We also focus on the clinical applicability of models to diseased anatomy, and the influence of their training and testing data. We examine publicly available datasets, computational demands, and evaluation metrics. Finally, we highlight the emerging research directions including multimodal integration and cross-modality frameworks. This review aims to provide researchers with a structured overview of current 3D reconstruction methodologies to identify opportunities for advancing deep learning towards more robust, generalizable, and clinically impactful solutions.
\end{abstract}


\end{frontmatter}



\section{Introduction}
Over the last few decades, magnetic resonance image (MRI)-based three-dimensional (3D) reconstruction has become essential in supporting medical diagnosis and clinical practice.  While clinicians are capable of interpreting 2D MR images and mentally constructing 3D anatomy, modern deep learning approaches aim to generate 3D models directly using machine learning. These accurate 3D representations of organs provide unparalleled insight into disease states and function, enabling more precise, effective, and personalized treatment for patients. Significant efforts have been made in statistical shape modeling (SSM) \citep{statisticalshapemodelAmbellan} and principal component analysis (PCA) \citep{PCAsurfacereconstruction2020} capturing general anatomical features. However, these models often fail to capture the subtle variations and complexities associated with individual patient pathologies. Accurately reconstructing the 3D shape of a patient’s anatomy from 2D MRI sequences remains a challenge, primarily due to the limited generalizability of current deep learning models and their tendency to struggle with variability in scan quality, anatomical diversity, and data scarcity. In a nutshell, 3D reconstruction for MRI aims to solve the problem of, given multiple 2D image stacks as input, recovering the 3D structure and geometry of the organ present in an area of interest. Mathematically, given $\mathcal{I} = \{I_1, I_2, \ldots, I_n\}$ as a set of 2D input images, the goal of 3D reconstruction is to infer a 3D shape $\hat{X}$ that approximates the unknown ground-truth shape $X$ of the object(s) present in the scan. Ideally, the reconstructed shape $\hat{X}$ should be as close as possible to $X$ in terms of geometry and structure.

\begin{figure}[!ht]
    \centering
    \includegraphics[width=1\linewidth]{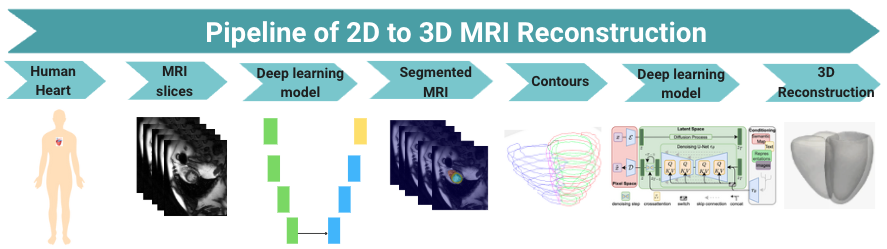}
    \caption{Example pipeline of 2D MRI to 3D reconstruction illustrating the process of image segmentation using U-Net \citep{optimizedxrayGopatoti2022}, then generating sparse contours as input into a diffusion model \citep{rombach2022highresolutionimagesynthesislatent} which can create a 3D mesh reconstruction of the heart \citep{qiao2025personalized}. All images used under Open Access licenses from their respective publications.}
    \label{fig:segmentation-3Dmodel}
\end{figure}

The rapid advancement of artificial intelligence and deep learning has revolutionized medical imaging, offering innovative solutions to long-standing challenges. Convolutional neural networks (CNNs)  \citep{RonnebergerUNET}, generative adversarial networks (GANs) \citep{FERREIRA2024103100, shende_brief_2019}, and diffusion models \citep{kazerouni_diffusion_2023} have demonstrated remarkable success in reconstructing high-quality 3D models from 2D MRI stacks. These deep learning architectures have been effectively applied across various anatomical structures from the heart to the brain to the femur, enabling detailed and anatomically accurate reconstructed organs. By leveraging large-scale datasets, these models often outperform traditional techniques, especially in scenarios complicated by motion artifacts or atypical pathologies. See Figure~\ref{fig:segmentation-3Dmodel} for an example pipeline of 2D-to-3D reconstruction. Consequently, the integration of deep learning into medical imaging pipelines not only makes 3D reconstructions more feasible, but also enhances their clinical relevance, paving the way for more accurate diagnosis and personalized treatment planning.

\citet{pandey_deep_2025}, \citet{deeplearningreconSamavati}, \citet{lin2025pixelspolygonssurveydeep}, \citet{oscanoa_deep_2023}, and \citet{webberDMmedicalreconstruction} present some of the most recent reviews on deep learning-based 3D reconstruction. \citet{pandey_deep_2025} offered a comprehensive overview of current challenges, network architectures, and datasets used specifically for 3D cardiac mesh reconstruction. A notable strength of their work is its discussion of the therapeutic implications of deep learning in the medical field. However, its exclusive focus on cardiac reconstruction limits exploration of techniques developed for other organs. In contrast, the review by \citet{deeplearningreconSamavati} provides valuable insight into general 3D reconstruction techniques using deep learning for multiple organs. The study covers both single-image and multi-image methods, offering details on point cloud and surface reconstruction, which have direct relevance to MRI applications. \citet{lin2025pixelspolygonssurveydeep} also expanded on other modalities such as computed tomography (CT), x-ray, ultrasound, and microscopy, providing a systematic review of image-to-mesh 3D reconstruction using deep learning. Their review is comprehensive in regard to methods and datasets but does not provide context on 3D modeling for personalized treatment, or the cross-modality integration of different models. \citet{oscanoa_deep_2023} focused solely on cardiac reconstruction, limiting cross-organ applicability, but stands out for its in-depth discussion of application-specific methods such as tissue characterization and blood flow estimation. The review by \citet{webberDMmedicalreconstruction} is specific to diffusion models, introducing the theory of diffusion model-based unsupervised reconstruction and identifying key research themes. They also address modality-specific challenges within MRI, CT, and ultrasound, and survey attempts to solve these challenges to image reconstruction with diffusion model.

In this review, we examine the latest models, methodologies, and challenges in applying deep learning to 3D MRI reconstruction across the human body, especially highlighting untapped opportunities where techniques developed for one organ may be transferable to others. We propose strategies to address current limitations and discuss how deep learning is shaping 3D reconstruction for personalized treatment and patient experience from a computational perspective. Finally, we assess the effectiveness of current evaluation metrics and offer recommendations to guide future research in this field. 

\section{Background}
\subsection{Search Strategy}
While this is not a systematic review, an extensive search was carried out for this review across multiple databases, including PubMed, Google Scholar, IEEE Xplore, SpringerLink, ScienceDirect, Wiley, ACM Digital Library, and arXiv Sanity Preserver in order to discover relevant material on the theory and application of deep learning algorithms for MRI reconstruction. We searched for  peer-reviewed journal papers or papers published in the proceedings of conferences or workshops, and preprints. Our customized search queries included ``3D reconstruction'', ``magnetic resonance'', ``magnetic resonance imaging'', ``deep learning'', and ``2D to 3D reconstruction''. We filtered our search results to remove false positives and only included papers that related to deep learning 3D reconstruction for MRI from 2D slices. We selected the papers for detailed examination based on careful evaluation of their novelty, contribution, and significance. Through the application of the aforementioned criteria, we aimed to provide a comprehensive overview of the most important and impactful studies.

\subsection{Challenges in 3D Deep Learning Reconstruction}
Despite notable advances in deep learning-based 3D reconstruction, several challenges remain, particularly when extending current methods to various organ systems. One core issue is the inconsistent voxel spacing and resolution across MRI datasets. Unlike CT, where voxel dimensions are generally standardized, MRI acquisitions often differ widely due to scanner-specific protocols and anisotropic slice thicknesses. This variation complicates spatial alignment, interpolation, and generalization across datasets \citep{AnatomicallyConstrainedNeuralNetworksOktay, chenchenimprovegeneralizability}. A lack of generalizability can further complicate clinical deployment. Models trained on data from a specific scanner, demographic, or imaging sequence often perform poorly when transferred to other settings without retraining or domain adaptation. These shifts are common in real-world multi-center workflows.

A further complication is the difficulty in modeling pathological anatomy. Many datasets are biased toward healthy or mildly abnormal cases, leading models to overfit to normal anatomical priors \citep{genderimbalancepnaspnas,biasindatasetsPuyol}. As a result, reconstructions of rare or complex pathologies—such as tumors, congenital malformations, or fibrotic lesions—may be inaccurate or over-smoothed \citep{factorizedspatialrepresentationAgisilaos}. This is particularly critical in organs with high inter-patient variability, such as the brain and heart.

Limited availability of large, annotated datasets also hinders progress. Annotating 3D medical images is time-consuming, resource-intensive, and prone to inter-observer variability \citep{oscanoa_deep_2023}. While some public datasets exist, they are often small in scale, lack full volumetric coverage, or are not representative of diverse patient populations \citep{pandey_deep_2025}. From a computational standpoint, training 3D convolutional networks or volumetric GANs demands significant GPU memory and processing power. This challenge is exacerbated when dealing with full-resolution 3D MRI volumes. Efficient architectures and compression strategies are being explored, but there remains a trade-off between model complexity, inference time, and reconstruction accuracy \citep{biswas_dynamic_2019}.

Finally, multimodal and multi-sequence integration remains an open problem. Different MRI sequences (e.g., T1, T2, FLAIR) provide complementary information, but vary in contrast, resolution, and noise characteristics. Aligning these volumes and extracting meaningful representations for 3D reconstruction is algorithmically and computationally challenging \citep{latentDMimagetoimagezhu}. Cine MRI typically acquires a set of 2D slices over a full cardiac cycle, which are sparse representations of the underlying 3D heart anatomy. This makes accurate 3D surface reconstruction difficult, especially in regions with little or no data.

\subsection{Reconstruction Methodologies and Architectures}

\begin{figure}[!ht]
    \centering
    \includegraphics[width=1.0\linewidth]{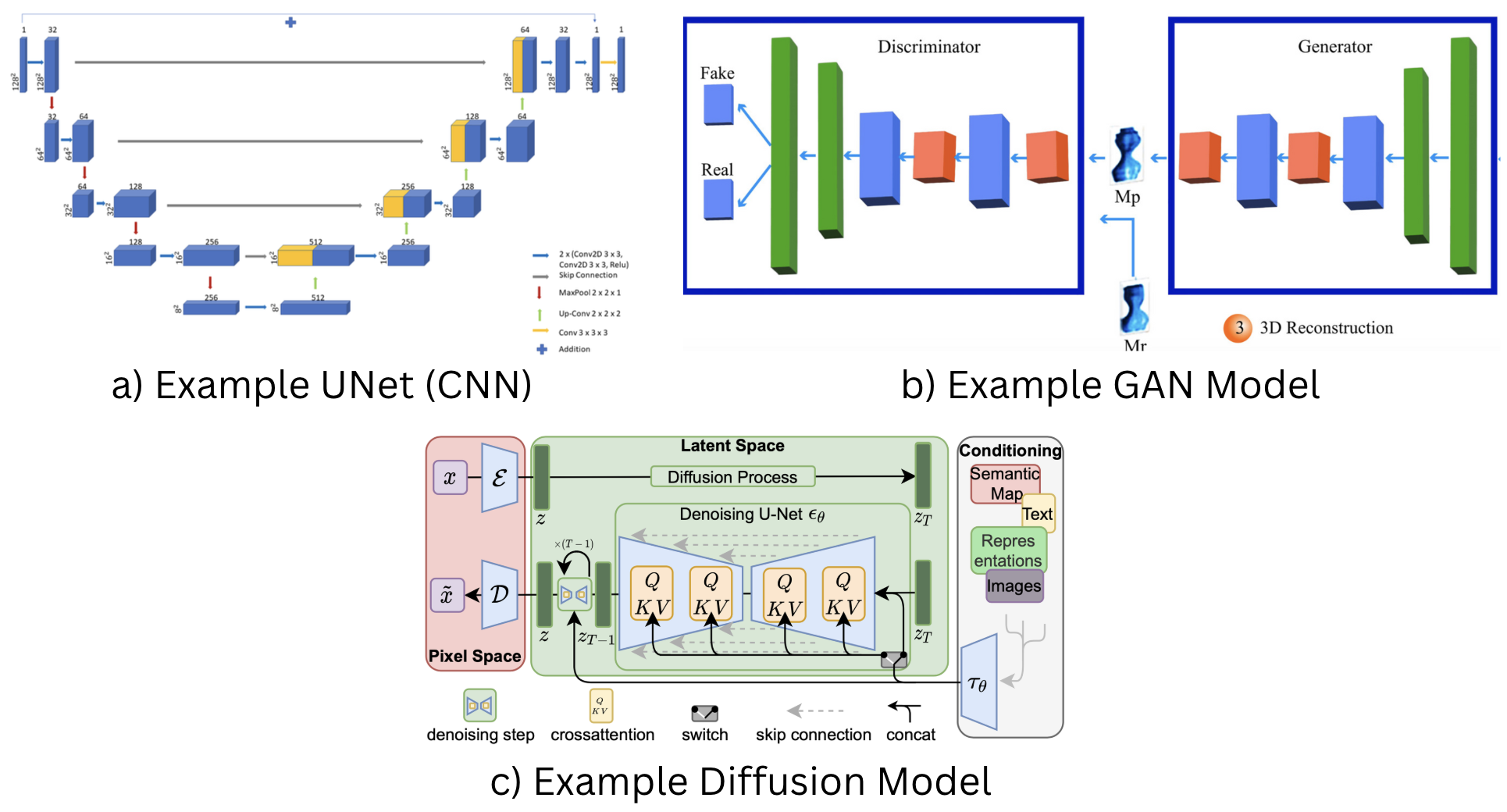}
    \caption{Example 3D reconstruction deep learning architectures. These models cannot be used for 3D reconstruction on their own, but can be used as part of a reconstruction pipeline. (a) Standard U-Net \citep{archie2021unet}. (b) General Adversarial Network (GAN) \citep{Alrashedys22114297} is an end-to-end pipeline that trains the generator in an adversarial manner to generate samples that the discriminator is capable of distinguishing from the real data sample. This example is GAN-LSTM wherein after extracting features by pretrained model, a 3D model of the tumour is reconstructed \citep{hong_gan-lstm-3d_nodate} (c) Diffusion models \citep{rombach2022highresolutionimagesynthesislatent} define a Markov chain of diffusion steps to slowly add random noise to data and then learn to reverse the diffusion process to construct desired data samples from the noise. All images used under Open Access licenses from their respective publications.}
    \label{fig:sample-dl-models}
\end{figure}

A wide range of deep learning methodologies have been explored for reconstructing 3D shapes from 2D slices, varying in data representations (e.g., voxel grid, mesh, point cloud, graph) and architectures such as CNNs \citep{RonnebergerUNET}, graph convolutional networks (GCNs) \citep{ReconstructingHigh-QualityDiffusionMRIHong}, GANs \citep{goodfellow_generative_2014}, diffusion models \citep{Croitoru_2023}, etc. See Figure~\ref{fig:sample-dl-models} for example architectures of a CNN, GAN, and diffusion model.

Traditional statistical shape-constrained methods such as Shape-Refined Multi-Task Learning and Anatomically Constrained Neural Networks (ACNNs) \citep{anatomicallyconstrainedoktay, brainACNN2024Arthy} incorporate prior anatomical knowledge into CNNs, either via auxiliary shape losses \citep{accuract3Dreconstructionwang2022} or embedded statistical models (as shown in Table~\ref{tab:method-comparison}). These approaches enforce plausible anatomical structures and improve robustness to noise and motion artifacts. However, their inductive bias toward normative anatomy reduces adaptability to out-of-distribution or pathological cases and increases dependency on well-annotated datasets. In contrast, Bayesian deep neural networks that integrate statistical shape modeling (e.g., Bayesian SSMs) model epistemic uncertainty, providing calibrated confidence estimates in ambiguous regions \citep{banerjee_completely_2021}. These methods are particularly useful when input slices are sparse or corrupted, but suffer from high training complexity and slow inference due to sampling-based posterior estimation.

\begin{table}[!ht]
\centering
\caption{Summary of 3D MRI Reconstruction Methodologies and Architectures.}
\begin{tabular}{|p{3cm}|p{2.5cm}|p{4.5cm}|p{4cm}|}
\hline
\textbf{Name} & \textbf{Category} & \textbf{Main Benefits} & \textbf{Main Drawbacks} \\
\hline
\textbf{Statistical Shape Models (SSM)} & Method & 
- Strong prior knowledge encoding \newline
- Interpretable and low-dimensional shape space \newline
- Good for limited data regimes &
- Limited flexibility \newline
- Poor generalization to unseen shapes \newline
- Hard to capture fine-grained or pathological variability \\
\hline
\textbf{Anatomically Constrained Neural Networks (ACNN)} & Method & 
- Incorporates anatomical priors and constraints \newline
- Improves clinical plausibility of reconstructions \newline
- Encourages shape consistency &
- Requires annotated priors or templates \newline
- May constrain model expressiveness \newline
- Less suitable for new anatomy \\
\hline
\textbf{Convolutional Neural Networks (CNN)} & Architecture & 
- Requires fewer parameters \newline
- Effective for volumetric and segmentation-based tasks \newline
- Compatible with dense 2D/3D data &
- Struggles with sparse or irregular input \newline
- Limited shape-awareness \newline
- Can overfit on small datasets \\
\hline
\textbf{Generative Adversarial Networks (GAN)} & Architecture & 
- Produces high-quality, realistic reconstructions \newline
- Good at capturing fine textures and details &
- Unstable training \newline
- Sensitive to mode collapse \newline
- Requires careful loss balancing \\
\hline
\textbf{Diffusion Models} & Architecture & 
- Realistic outputs and high resolution \newline
- Generates smooth and diverse outputs \newline
- Robust to sparse or noisy input &
- High computational cost \newline
- Requires long inference time due to iterative sampling \\
\hline
\end{tabular}
\label{tab:method-comparison}
\end{table}

\begin{figure}[!ht]
    \centering
    \includegraphics[width=1.0\textwidth]{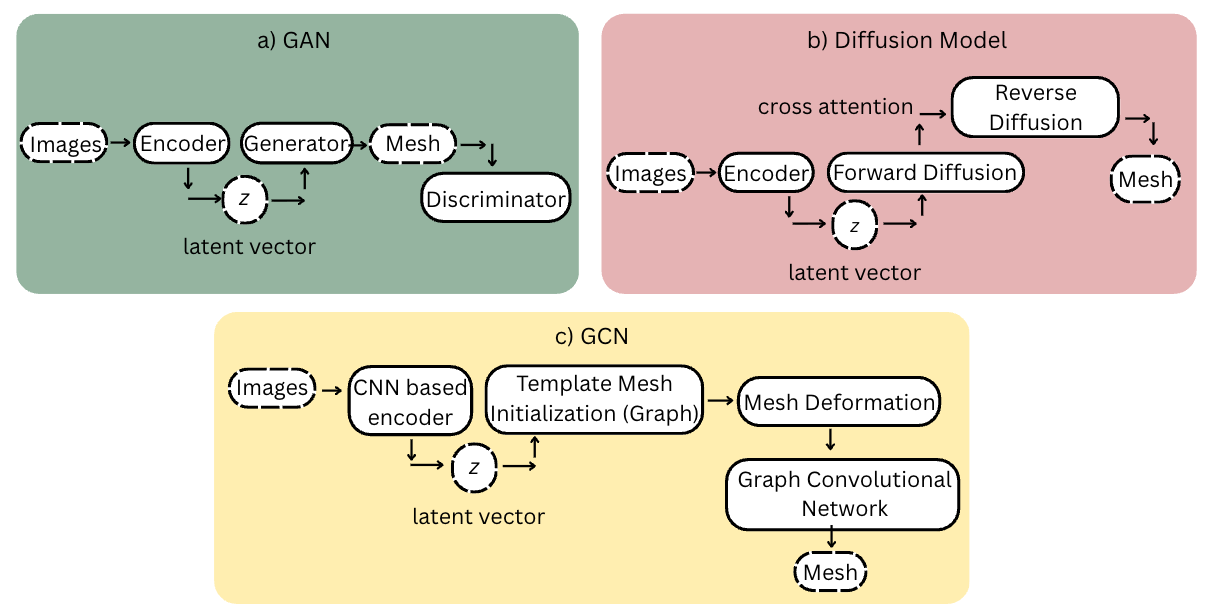}
    \caption{Diagrams for 3D mesh generation. a) Schematic diagram of a generative adversarial network (GAN)-based model for mesh generation. 2D images are inputted into the encoder which produces a latent vector. The latent vector is then passed to the generator which generates a point cloud. The point cloud is inputted into a discriminator network which differentiates real from fake shapes. b) Schematic diagram of a diffusion model for 3D mesh generation. Diffusion models gradually transform noise into structure. A feature encoder processes the images to produce a latent vector or embedding that summarizes the organ’s 3D shape. c) Schematic diagram of a graph convolutional network (GCN). The GCN deforms a template mesh using encoded features and local graph operations. Mesh losses enforce smoothness, realism, and anatomical correctness. }
    \label{fig:diagrams-dl-mesh-methods}
\end{figure}

Non-Euclidean architectures have also been adopted to better model surface and relational priors. Mesh-based methods such as MeshCNN \citep{MeshCNNHanocka} and GCN-based direct mesh reconstruction \citep{zhang2025reconsidertemplatemeshdeep} directly operate on triangular meshes, preserving geometric consistency and allowing end-to-end mesh generation. These models, however, often require pre-defined mesh topology and complex preprocessing pipelines. Point cloud-based networks such as Point2Mesh-Net \citep{beetz_point2mesh-net_2022} introduce hierarchical feature aggregation, improving the reconstruction of localized features while still capturing global features. Still, these models struggle to capture fine-grained structure without high point density or specialized sampling strategies. Graph neural networks (GNNs), including spectral GCNs and Graph U-Nets \citep{ReconstructingHigh-QualityDiffusionMRIHong} (see Figure~\ref{fig:diagrams-dl-mesh-methods}), have been proposed for learning over anatomical structures represented as graphs. These methods exploit topological relationships and are well-suited for semi-supervised learning or population-level analysis. However, they assume static graph structures and may lose local detail through pooling operations.

GANs \citep{mehri_generative_2023, shende_brief_2019} have been used to augment datasets via synthetic 2D slices or 3D volumes. Although effective in boosting performance, GANs often suffer from mode collapse and unstable training. More recently, diffusion models have emerged as state-of-the-art generative models for medical imaging \citep{he2023dmcvrmorphologyguideddiffusionmodel, lee_improving_2023}. These models learn the reverse of a noise corruption process, yielding high-fidelity and diverse samples. Their potential for 3D MRI reconstruction lies in their stability, controllability, and ability to model complex data distributions. Recent adaptations of diffusion frameworks for 3D data (e.g., volumetric DDPMs \citep{wu2024unpairedvolumetricharmonizationbrain}) show promise in synthesizing anatomically realistic volumes even from sparse or anisotropic inputs, though they remain computationally expensive and require further optimization for clinical deployment.

While CNN-based volumetric methods remain dominant due to their simplicity and effectiveness, mesh-, point cloud-, and graph-based models offer improved geometric expressiveness. Generative models, particularly diffusion-based ones, are likely to play a growing role in data augmentation and direct reconstruction as their scalability improves. We review the application of these techniques in subsequent sections.

\section{Methods and Models}
In the following subsections, we provide a detailed methodological analysis of how deep learning addresses the challenges inherent in traditional 3D MRI reconstruction. We examine the latest models and techniques, and explore potential avenues for further improvement of 3D reconstruction. \hbox{Table~\ref{tab:research_papers}} provides a summary of all reviewed 3D reconstruction papers, models, and datasets.

\subsection{Point Cloud-Based Reconstruction}
Point clouds \citep{pointcloudishereRusu} are a flexible geometric representation of anatomical structures where each point corresponds to a specific location in 3D space. Unlike voxel grids or meshes, point clouds are unordered sets of spatial coordinates. They can optionally include intensity or feature values, which capture the surface or internal features of an organ without imposing a fixed grid or connectivity.

\begin{figure}[!ht]
    \centering
    \includegraphics[width=0.8\linewidth]{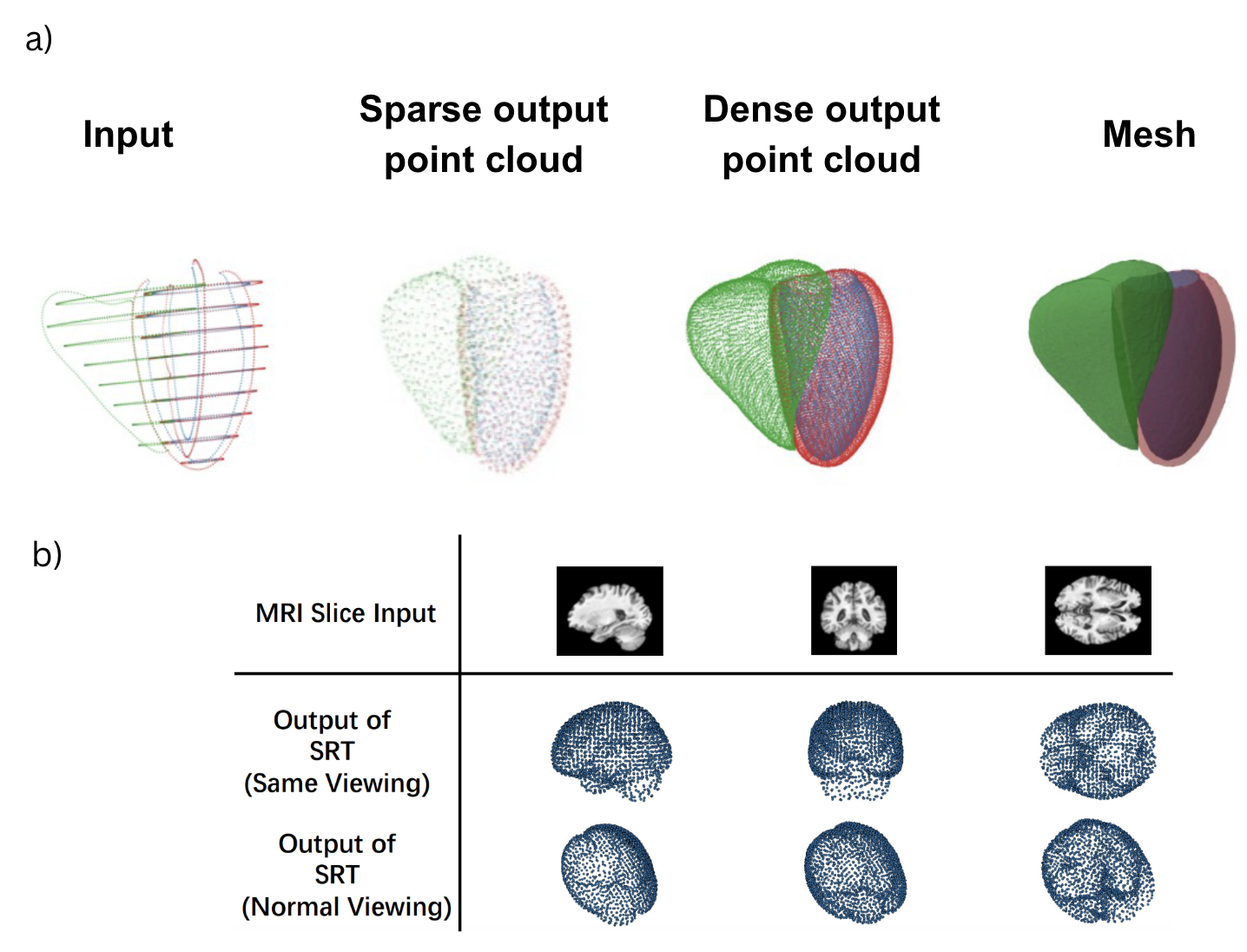}
    \caption{Examples of 3D MRI point cloud reconstructions. a) Dense point cloud and output mesh created by PCCN of the human heart ventricles \citep{beetz_multi-class_2023}. b) Different angle MRI scans and the generated point clouds from \citep{SRThu2022}. All images used under Open Access licenses from their respective publications.}
    \label{fig:PC-methods}
\end{figure}

One successful example of a lightweight point cloud reconstruction pipeline is the Multi-Class Point Cloud Completion Network (PCCN) \citep{beetz_multi-class_2023}, which reconstructs biventricular cardiac anatomy from sparse, misaligned point sets derived from cine-MRI contours (Figure~\ref{fig:PC-methods}a). To tackle sparsity in 2D MRI images, PCCN utilizes an encoder-decoder architecture which takes a two-fold approach. First, a coarse, sparse low-density point cloud is produced to capture the global surface structure. Second, a dense, high-resolution point cloud is generated by deforming grid-structured patches, enhancing local anatomy. PCCN outperforms conventional voxel-based models, such as 3D U-Net, with significantly lower Hausdorff \citep{HDtran_chapter_2016} and Chamfer distances \citep{lin2024hyperbolicchamferdistancepoint} across cardiac classes. Although its training and testing exclusively on healthy patients remains a limiting factor, this motivates future work for generalizability to diseased patients.

Many approaches have exploited multi-view MRI data to enhance spatial completeness. \citet{automatic3DsurfacereconstructionLAXiong} used a CNN-based regression network that maps irregular point clouds onto dense 3D surface meshes without relying on intermediate volumetric or voxel representations. A 30-layer 3D fully CNN backbone was designed to generate 3D left atrium (LA) reconstructions directly from sparse 3D point clouds. The pipeline starts with 3D point clouds, converts them into a dense 3D volumetric image, and then uses a 3D CNN to classify each voxel within that volume, effectively generating a 3D surface reconstruction of the LA. While promising, the method is directly dependent on the coverage of the point clouds, meaning low coverage point clouds will produce low-quality meshes. Introducing anatomical constraints to the loss function to ensure that the outputs contain all key anatomical landmarks would be highly beneficial in clinical applications. An in-depth overview of loss functions and evaluation metrics can be found in the Appendix.

Beyond the heart, point cloud methods have also been extended to neuroimaging. A recent study \citep{SRThu2022} developed a CNN-free advanced spatial registration transformer (SRT) network based on the self-attentive mechanism to align 2D-derived brain point clouds into a canonical space (Figure~\ref{fig:PC-methods}b). The SRT utilizes an encoder based on a multi-headed attention module (MAM) \citep{multiheadattention2021}. The encoder extracts image features from the input 2D MRI and outputs a feature vector that obeys a normal distribution of specific mean and variance. Then a decoder, also composed of multiple MAMs, forms an ``up-down-up'' generative system to reconstruct the 3D point cloud from this feature vector and outputs an accurate point cloud of 2,048 points. While the model achieved promising scores in chamfer distance, and earth mover’s distance, the SRT does not take into account pathological conditions of the MRI scans and may not be applicable to MRI's from different sites and scanners.

As demonstrated in cardiac and neuroimaging applications, point cloud methods excel at capturing global anatomical structures and enabling fast inference. However, these pipelines often rely on healthy, well-covered anatomical regions, limiting their applicability to pathological or incomplete data. A recurring theme in many works is the use of point clouds as an intermediate representation for surface reconstruction. In the following subsection, we expand on this idea, highlighting how deep learning frameworks have been adapted for direct mesh generation from both imaging and point cloud data.

\subsection{Mesh-Based Reconstruction}
Deep learning models have been developed to directly regress mesh vertices from image features, bypassing intermediate voxel or point cloud stages entirely. These architectures are optimized using mesh-specific losses (e.g., edge length or surface Laplacian) to preserve anatomical shape and surface smoothness. Mesh-based 3D reconstruction involves generating a continuous surface representation of anatomical structures using interconnected vertices, edges, and faces—typically forming a polygonal mesh. Unlike point cloud methods, which represent geometry as unordered sets of discrete points, mesh-based techniques explicitly capture surface topological information, allowing for detailed and anatomically coherent reconstructions. This structured representation is particularly advantageous for downstream tasks such as simulation, visualization, or quantitative analysis, where preserving surface continuity is essential.

\begin{figure}[!t]
    \centering
    \includegraphics[width=0.6\linewidth]{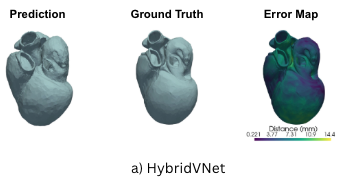}
    \caption{Exemplar of deep learning mesh-based 3D reconstructions. a) HybridVNet \citep{HybridVNetGaggion_2025}. This image is used under Open Access licenses from their respective publications.}
    \label{fig:3D-cardiac-mesh-reconstructions}
\end{figure}

In cardiovascular imaging, combining sparse point clouds with mesh prediction has shown particular promise. A recent approach called Point2Mesh-Net \citep{beetz_point2mesh-net_2022} leverages sparse 3D landmarks to infer high-resolution cardiac meshes by deforming a template mesh through a learned point cloud-to-mesh regression network (Figure~\ref{fig:3D-cardiac-mesh-reconstructions}a).
Point2Mesh-Net enables multi-scale feature learning and helps the network overcome challenges like data sparsity and slice misalignment.
The network's encoder inputs learns multi-scale features from sparse, misaligned contours represented as 900-point 3D point clouds. Its decoder gradually maps latent vectors to high-resolution 3D triangular meshes with 1,780 consistently connected vertices, overcoming data sparsity and misalignment issues. Complementary work by \citet{chen_shape_2021} demonstrates robust shape registration from point clouds by learning deformation fields that morph a generic template mesh onto a target anatomy using a smoothness-constrained loss function. These methods highlight the utility of mesh templates for encoding anatomical priors and ensuring consistent topology.

Another promising direction involves hybrid architectures that combine volumetric and surface-based representations. HybridVNet \citep{HybridVNetGaggion_2025} is a recent GCN that integrates a hybrid variational encoder–decoder, enabling end-to-end training on cardiac MRI while predicting detailed 3D surface meshes (Figure~\ref{fig:3D-cardiac-mesh-reconstructions}b). The model employs a dual-branch design: the first branch (an image convolutional encoder) learns a latent representation of the input images, while the second branch (a spectral graph convolutional decoder) generates a graph representation. This hybrid formulation allows the model to learn from both spectral intensity patterns and surface-based priors, leading to improved anatomical accuracy and mesh consistency compared to voxel-only or mesh-only pipelines.
\citet{verhulsdonk_shape_2024} developed Cardiac DeepSDF, which creates patient-specific 3D heart meshes by integrating signed distance functions with a Lipschitz regularization term in order to learn a comprehensive representation of cardiac anatomy within a continuous function space. Unlike linear statistical models, this method allows for advanced topological changes in cardiac images. While Cardiac DeepSDF has only been applied so far to encode cardiac biventricular surfaces, the model can be potentially extended to encode additional chambers like the atria or the aorta, which would increase its clinical relevance.

\citet{pmlr-v194-beetz22a}, employs a mesh deformation U-Net which converts contours into a point cloud, fits a template mesh, and finally applies the U-Net to correct slice misalignment. In the template mesh fitting stage, the template mesh from an SSM dataset \citep{bai2015bi} is approximately aligned with the sparse point cloud by minimizing the earth mover's distance. This introduces an anatomical shape prior and converts sparse point clouds into dense meshes while preserving consistent vertex connectivity across the dataset. The fitted mesh, which may still be misaligned, is then refined by the pre-trained mesh deformation U-Net. Leveraging multi-view MRI information and the anatomical prior, the U-Net corrects for motion-induced slice misalignment.

Together, these methods demonstrate the growing maturity and diversity of mesh-based reconstruction techniques. From hybrid volumetric-mesh networks to sparse-data deformation models, the field is rapidly evolving toward anatomically accurate, topologically consistent surface models suitable for both individualized clinical use and large-scale population analysis. Future work is likely to focus on improved generalization across pathologies, integration with multimodal inputs, and the development of clinically validated, explainable reconstruction frameworks, as it will be discussed in later sections.

\subsection{Shape Aware Reconstruction Methodologies}
Robust shape registration techniques also play a critical role in mesh-based modeling. By leveraging statistical or learned priors such as anatomical templates, latent shape manifolds, or population-level atlases, these models constrain the output space and prevent unrealistic deformations using pre-defined priors as guides for template meshes and point clouds. Shape aware reconstruction can be a crucial step in ensuring anatomical plausibility, especially when input data is sparse, noisy, or only partially covers the target structure, although it can limit generalizability among different populations of test data.

\begin{figure}[!ht]
    \centering
    \includegraphics[width=0.6\linewidth]{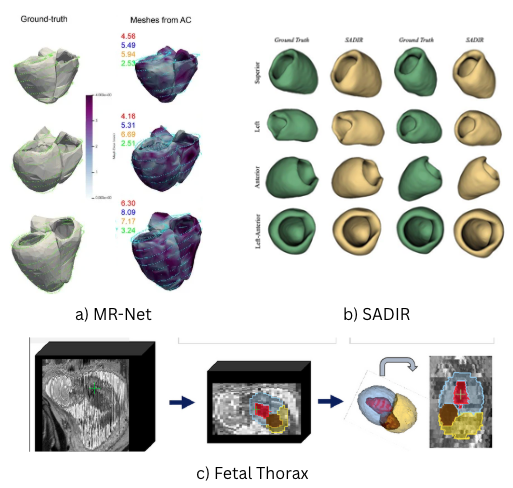}
    \caption{Exemplars of shape aware deep learning 3D reconstructions. a) Sample of 3D cardiac shape reconstruction from MR-Net \citep{HybridVNetGaggion_2025} using automatic annotated contours. Each row is one sample (left: ground-truths; right: reconstructed 3D meshes). b) 3D myocardium reconstructed from sparse 2D slices by the shape aware diffusion model (SADIR) over four different views \citep{jayakumar_sadir_2023}. c) Reconstruction of the fetal thorax from motion-corrupted MRI stacks \citep{automatedfetalreconstructionUus2022}. All images used under Open Access licenses from their respective publications.}
    \label{fig:shape-aware-examples}
\end{figure}

\citet{chen_shape_2021} proposed a hybrid GCN for 3D mesh reconstruction, MR-Net, that learns continuous deformation fields to warp a canonical cardiac template mesh onto subject-specific anatomy represented as point clouds (Figure~\ref{fig:shape-aware-examples}a). The deformation module gradually deforms a template mesh to the personalized target mesh, preserving the topology and connectivity of the meshes. It includes three GCN blocks, each comprising 14-15 graph convolution layers. Each GCN block predicts an output of the target mesh, allowing the template mesh to be deformed progressively to fit the contours. This registration process is driven by a multi-term mesh loss function that balances geometric alignment with smoothness constraints, ensuring that the resulting meshes are both anatomically accurate and topologically consistent.

The shape aware diffusion model (SADIR) \citep{jayakumar_sadir_2023} has been shown to learn brain and cardiac shape priors using an atlas building framework (Figure~\ref{fig:shape-aware-examples}b). SADIR works in two parts: an atlas building network parameterized to provide a mean output image, and a reconstruction network that considers each reconstructed image as a deformed variant of the obtained atlas. SADIR's use of this dual framework and prior information resulted in little difference from the ground truth in terms of anatomical structure, even without taking voxel intensity into account.
Other methods use atlases to correct for motion artifacts. \citet{automatedfetalreconstructionUus2022} proposed an automated motion correction pipeline for reconstruction of 3D fetal thorax anatomy from fetal MRI stacks (Figure~\ref{fig:shape-aware-examples}c). The pipeline has four main parts: a standard radiological atlas space, a fetal trunk atlas for landmark definition, a CNN training method with atlas-based masks, and the generation of a common template space. The standard atlas space acts as a common reference for reorienting all input MRI stacks. The fetal trunk atlas references point landmarks, defined in an average fetal trunk atlas that is reoriented to the standard radiological space. The adaptive common template space accounts for stacks affected by low image quality or extreme motion and provides a stable initial registration target. The common space also accounts for inter-subject deviations by not relying solely on direct registration to a fixed atlas.

Shape aware anatomical priors can be embedded directly into loss functions, used to initialize deformation fields, or integrated into generative models. By aligning reconstruction tasks with known anatomical variability, shape-aware frameworks can enhance generalizability across subjects and improve robustness in clinically diverse datasets. However, they can also be a hindrance in cases of extremely deformed organs not fitting with expected healthy models.

\subsection{Volumetric Reconstruction Methodologies}
\begin{figure}[!ht]
    \centering
    \includegraphics[width=1.0\linewidth]{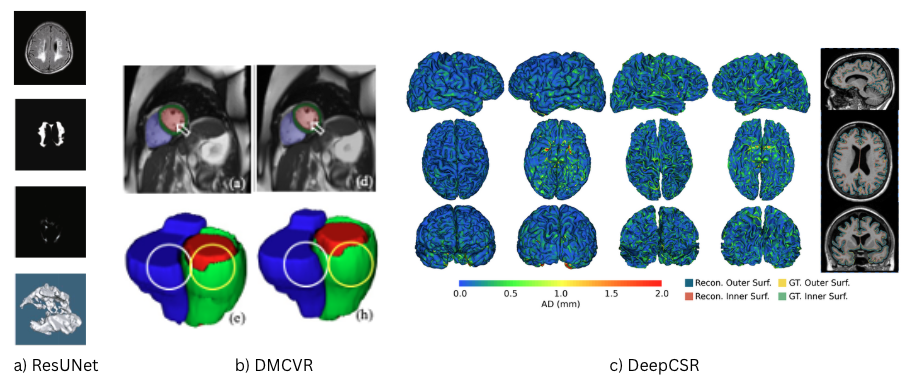}
    \caption{Exemplars of deep learning volumetric 3D reconstructions. a) Segmentation and reconstruction renderings of a group of white matter hyperintensity images with 3D reconstruction results obtained from the segmented image \citep{accuract3Dreconstructionwang2022}. b) 2D and 3D visualization results of DMCVR. The generated cardiac segmentation and reconstruction of original image (left column) and DMCVR result (right column) \citep{he2023dmcvrmorphologyguideddiffusionmodel}. c) Outer and inner cortical surfaces reconstructed with DeepCSR \citep{cruz2020deepcsr3ddeeplearning}. All images used under Open Access licenses from their respective publications.}
    \label{fig:volumetric-reconstruction-examples}
\end{figure}

Volumetric reconstruction methodologies involve generating a full 3D voxel-based representation, where 3D shapes become parameterized as grids of voxels. See Figure~\ref{fig:volumetric-reconstruction-examples} for visualized volumetric reconstructions.

The regularized 3D diffusion model (R3DM) \citep{bangun2024mrireconstructionregularized3d} attempts a hybrid approach that combines data-driven priors and model-based reconstruction techniques. R3DM integrates a 2D U-Net with cross-attention on the third dimension with an optimization method for knee MRI 3D reconstruction. R3DM generates samples with DDPM \citep{ho2020denoisingdiffusionprobabilisticmodels} by iteratively removing noise, alternating stochastic sampling steps with optimization updates. This iterative process refines the generated images by continually incorporating both learned data distribution and measurement constraints with general prior knowledge. The authors incorporate $k$-space measurements as a guiding constraint within the diffusion model's sampling process, which helps with enhancing image quality. By combining these robust diffusion models with conventional optimization, R3DM achieves good performance in out-of-distribution evaluation; but due to its use of a pre-trained diffusion model, it has a dependency on high-quality input images.

\citet{accuract3Dreconstructionwang2022} developed two CNNs, ResUnet and an attention U-Net, for accurate 3D reconstruction of white matter hyperintensities (WMH) in the brain, a condition linked to cognitive decline and dementia. ResUnet is an extension of ResNet~\citep{He2015DeepRLresnet} and enhances the feature extraction of ResNet. ResUnet addresses network degradation by adding residual modules and incorporating jump connections to the original ResNet architecture, demonstrating better segmentation accuracy than ResNet. The Attention-UNet, developed by the same authors, sequentially applies channel and spatial attention modules, so that each node can learn `what' and `where' to attend in the channel and spatial axes, respectively. Instead of the classic marching cubes algorithm \citep{marchingcubeslorensen}, which can suffer from ambiguity problems in connecting triangular surfaces, the ResUnet and Attention-UNet employ the moving tetrahedron method for 3D reconstruction. This method divides the cube element into tetrahedra, which helps resolve ambiguity and achieve better modeling accuracy. The Attention-UNet achieves higher segmentation accuracy than ResUnet and lays an important foundation for accurate 3D reconstruction (Figure~\ref{fig:volumetric-reconstruction-examples}a). This work is the first attempt to carry out accurate 3D reconstruction of WMH, offering novel ideas for quantitative analysis and reconstruction.

DiffusionMBIR \citep{chung_solving_2022} is a novel method for 3D medical image reconstruction that addresses the computational challenges of applying 2D diffusion models to higher dimensions. DiffusionMBIR achieves high-quality reconstructions by using 2D diffusion models that process slices, combined with a 3D model-based iterative reconstruction approach to ensure 3D coherence. It combines the strengths of pre-trained 2D diffusion models with model-based iterative reconstruction (MBIR), using a total variation (TV) prior along the z-axis to ensure inter-slice coherence. To achieve coherent 3D reconstructions, DiffusionMBIR augments the 2D diffusion prior with a model-based prior. This model-based prior enforces spatial correlation across slices, addressing the issue of incoherence that would arise if 2D slices were reconstructed independently. The model runs the diffusion-based denoising step slice-by-slice, while a 3D alternating direction method of multipliers (ADMM) \citep{admmmultiplier2011} update step is used to impose data consistency with the z-directional TV prior across the entire volume.

DMCVR \citep{he2023dmcvrmorphologyguideddiffusionmodel} is a novel morphology-guided diffusion model designed for 3D cardiac volume reconstruction from sparse 2D cine MRI (Figure~\ref{fig:volumetric-reconstruction-examples}b). DVCMR creates 3D reconstructions without generating meshes by leveraging global, stochastic, and regional morphology latent codes to guide the generative process. The global semantic encoder captures general high-level semantic features, while the regional encoder embeds the 2D image into a latent space that contains the necessary information to produce a segmentation map of the target cardiac tissues. The stochastic latent code embeddings are interpolated over a unit sphere. The stochastic latent space itself does not contain interpolable high-level semantics. The forward diffusion process takes the noise $x_T$ as input and produces $x_0$
the target image. Since the change in $x_T$ will affect the details of the output
images, we can treat $x_T$ as the stochastic latent code. Specifically, the regional encoder represents the shapes and locations of the left ventricle cavity, left ventricle myocardium, and right ventricle cavity. To generate missing slices for a smooth super-resolution cine image volume, DMCVR interpolates additional linear and spherical interpolation latent codes. DMCVR was shown to outperform other methods such as DeepRecon \citep{deepreconQi2022}, a GAN based interpolation network which uses latent space to interpolate the missing information between
adjacent 2D slices, in both Dice score and Hausdorff distance.
\citet{Zhangs21092978} also used interpolation to reconstruct high-resolution 3D brain MRI from lower-resolution 2D slices. Their method leverages two specialized GANs--a receptive field block enhanced super-resolution GAN and a noise-based super-resolution GAN--to enhance image detail and fill in missing information during the reconstruction process. The model effectively replaces computationally expensive 3D CNNs using this two-pronged approach.

DeepCSR \citep{cruz2020deepcsr3ddeeplearning} provides a deep learning framework designed for cortical surface reconstruction  (Figure~\ref{fig:volumetric-reconstruction-examples}c). The model predicts implicit surface representations for points in a continuous coordinate system, enabling the reconstruction of high-resolution surfaces that capture fine details of cortical folding. Surfaces are implicitly defined as the level set of a continuous function that maps points in 3D Euclidean space to a scalar using a neural network model with hypercolumn features. Input point coordinates are projected into each feature map, and feature values are linearly interpolated at these locations. Another method by \citet{JUREK2020111} reconstructs blood vessel tree images by developing a CNN that processes information from multiple orthogonal thick-slice scans, leveraging a super-resolution CNN by incorporating information from multiple orthogonal brain MRI scan views. Instead of using a single input, this CNN is driven by patches taken from three orthogonal thick-slice images (axial, coronal, and sagittal scans). While this is a promising method for small structure reconstruction, it was shown to be highly dependent on image content and orientation. The authors hence suggested that learning more complex relationships within the data might further improve future reconstructions.

\begin{longtable}{p{4.35cm}p{2.15cm}p{3.5cm}p{2.7cm}}
    \caption{Summary of deep learning models for 3D MRI reconstruction.} 
    \label{tab:research_papers} \\
    \hline
    \textbf{Paper Title} & \textbf{Model Type} & \textbf{Dataset (Organ)} & \textbf{Model Name} \\
    \hline
    \endfirsthead

    \hline
    \textbf{Paper Title} & \textbf{Model Type} & \textbf{Dataset (Organ)} & \textbf{Model Name} \\
    \hline
    \endhead

    Shape registration with learned deformations for 3D shape reconstruction \citep{chen_shape_2021} & CNN & UK Biobank (heart) \citep{bycroft2018uk} & MR-Net \\
    \hline
    CNN-based superresolution reconstruction of 3D MR images \citep{JUREK2020111} & CNN & BrainWeb SBD (brain) \citep{cocosco1997brainweb} & Not specified \\
    \hline
    Reconstruction of 3D knee MRI using deep learning and compressed sensing \citep{dratsch_reconstruction_2024} & CNN & Proprietary (knee) & CS-AI \\
    \hline
    Deep-learning-reconstructed high-resolution 3D cervical spine MRI \citep{jardon_deep-learning-reconstructed_2023} & CNN & Proprietary (spine) & 3D-DLRecon \\
    \hline
    Multi-class point cloud completion networks for 3D cardiac anatomy reconstruction \citep{beetz_multi-class_2023} & CNN & UK Biobank (heart) \citep{bycroft2018uk} & PCCN \\
    \hline
    CNN-based superresolution reconstruction of 3D MR images using thick-slice scans \citep{JUREK2020111} & CNN & BrainWeb SBD (brain) \citep{cocosco1997brainweb} & SRCNN \\
    \hline
    A deep-learning approach for direct whole-heart mesh reconstruction \citep{kong_deep-learning_2021} & CNN & Multiple MRI (heart): MMWHS \citep{zhuang2018multivariate}, SLAWT \citep{karim2018algorithms}, LASC \citep{tobon2015benchmark} & 3D UNet \\
    \hline
    LaMoD: Latent motion diffusion model for myocardial strain generation \citep{xing_lamod_2025} & Diffusion Model & Multiple, DENSE, Cine MRI (heart) & LaMoD \\
    \hline
    SADIR: Shape-aware diffusion models for 3D image reconstruction \citep{jayakumar_sadir_2023} & Diffusion Model & OASIS-III (brain) \citep{lamontagne2019oasis}, MedShapeNet (heart) \citep{li2025medshapenet} & SADIR \\
    \hline
    Improving 3D imaging with pre-trained perpendicular 2D diffusion models \citep{lee_improving_2023} & Diffusion Model & IRB-approved in-house MRI (brain) & TPDM \\
    \hline
    DMCVR: Morphology-guided diffusion model for 3D cardiac volume reconstruction \citep{he2023dmcvrmorphologyguideddiffusionmodel} & Diffusion Model & UK Biobank (heart) \citep{bycroft2018uk} &  DMCVR\\
    \hline
     Solving 3D inverse problems using pre-trained 2D diffusion models \citep{chung_solving_2022} & Diffusion Model & BRATS (brain) \citep{menze2014multimodal} & DiffusionMBIR \\
     \hline
     3D MRI synthesis with slice-based latent diffusion models: improving tumor segmentation tasks in data-scarce regimes \citep{kebaili_3d_2024} & Diffusion Model & BRATS (brain) \citep{menze2014multimodal} & SBLDM \\
     \hline
     Diffusion probabilistic models for 3D point cloud generation \citep{ho2020denoisingdiffusionprobabilisticmodels} & Diffusion Model & ShapeNet (various) \citep{chang2015shapenet} & Not specified \\
     \hline
     Generation of 3D brain MRI using auto-encoding generative adversarial networks \citep{kwon_generation_2019} & GAN & ATLAS (brain) \citep{liew2018large}, BRATS (brain) \citep{menze2014multimodal} & 3D-$\alpha$-WGAN-GP \\
     \hline
     3D MRI reconstruction based on 2D generative adversarial network super-resolution \citep{Zhangs21092978} & GAN & IXI database (brain) \citep{rowland2004information} & RFB-ESRGAN \\
     \hline
     GAN-based motion artifact correction of 3D MR volumes using an image-to-image translation algorithm \citep{reddy_gan-based_2024} & GAN & ADHD-200 (brain) \citep{bellec2017neuro} & Not specified \\
     \hline
    Neural deformable models for 3D bi-ventricular heart shape reconstruction \citep{ye_neural_2024} & Neural Deformable Model & Cine MRI (heart) \citep{bai2015bi} & NDM \\
    \hline
    Point2Mesh-Net: Combining point cloud and mesh-based deep learning for cardiac shape reconstruction \citep{beetz_point2mesh-net_2022} & GCN & UK Biobank (heart) \citep{bycroft2018uk} & Point2Mesh-Net \\
    \hline
    End-to-End 4D Heart Mesh Recovery Across Full-Stack and Sparse Cardiac MRI \citep{chen2025end} & GCN & ACDC, M\&Ms, M\&Ms-2 (heart) \citep{bernard2018deep, campello2021multi, martin2023deep} & TetHeart \\
    \hline

\end{longtable}

\section{Non MRI Reconstruction Methods and Potential Applications}
\begin{figure}[!ht]
    \centering
    \includegraphics[width=1.0\linewidth]{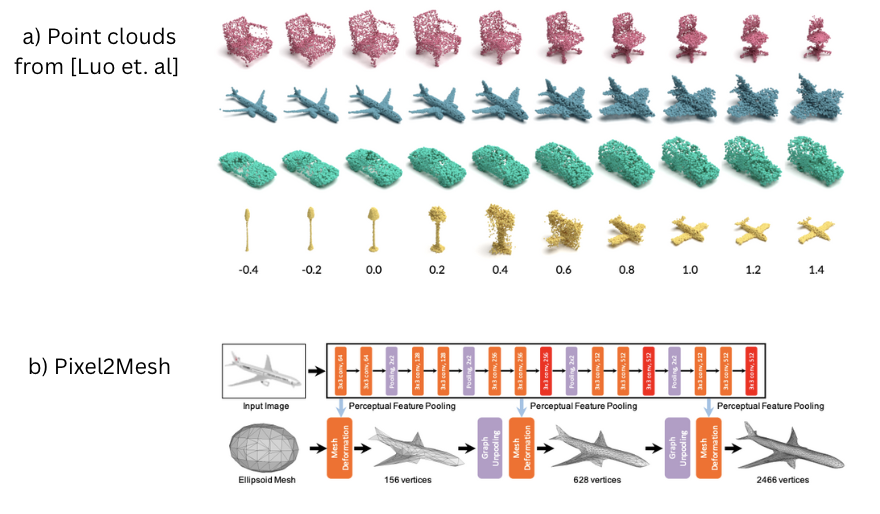}
    \caption{Non MRI outputs for 3D reconstruction. a) Latent space interpolation and extrapolation from the probabilistic diffusion model in \citep{luo2021diffusionprobabilisticmodels3d}. b) The cascaded mesh deformation network from Pixel2MeshNet \citep{wang_pixel2mesh_2018}. All images used under Open Access licenses from their respective publications.}
    \label{fig:non-mri-reconstructions}
\end{figure}

While the majority of 3D reconstruction methods discussed thus far were developed specifically for MRI data, there is a growing body of work in adjacent domains, such as animation and even natural image processing, that offers novel architectures and strategies with strong potential for translation to MRI-based workflows. These approaches often address challenges analogous to those in MRI reconstruction, including sparse sampling, resolution non-uniformity, and the need for anatomical priors. This section highlights innovative methods that could be adapted or extended to improve MRI-based 3D reconstruction. See Figure~\ref{fig:non-mri-reconstructions} for visualized non-MRI reconstructions.

Inspired by thermodynamic diffusion processes, \citet{luo2021diffusionprobabilisticmodels3d} proposed viewing points as particles that diffuse from an original distribution to a noise distribution for point cloud generation (Figure~\ref{fig:non-mri-reconstructions}a). The key novelty comes from learning the reverse diffusion process, which transforms this noise back into a desired shape, modeled as a Markov chain conditioned on a latent shape. This approach offers advantages over previous generative models for point clouds, as it does not rely on complex adversarial losses, require invertibility like flow-based models, or assume a generation ordering unlike auto-regressive models. This can be a particularly helpful technique for generalizing 3D point clouds to more diverse anatomical structures, especially in settings where MRI-derived point clouds are sparse, noisy, or exhibit high inter-subject variability, such as the human brain or heart. By modeling the shape distribution in a probabilistic and iterative fashion, this method could improve the robustness and fidelity of MRI-based 3D reconstructions, particularly when paired with anatomical priors \citep{biswas_dynamic_2019} or used to augment training data in low-resource domains \citep{chenchenimprovegeneralizability}.

In the field of ultrasound imaging, \citet{3dultrasoundreconstructioneid2025} proposed RapidVol, a novel framework for fast and accurate 3D reconstruction from 2D freehand ultrasound scans. RapidVol employs a hybrid implicit-explicit representation based on tri-plane decomposition and a lightweight multi-layer perceptron (MLP), enabling efficient reconstruction of structures such as the fetal brain. Its key innovation lies in its compact representation of the 3D volume: instead of storing a dense 4D tensor $V' \in \mathbb{R}^{H \times W \times D \times F}$ (where $F$ denotes feature channels), RapidVol represents the volume as $F$ sets of three 2D planes. This reduces the memory complexity from $\mathcal{O}(n^3)$ to $\mathcal{O}(n^2)$, cutting storage requirements from 1049\,MB to just 15\,MB on their dataset. Bilinear interpolation is applied within each tri-plane to handle discontinuities and evaluate features at non-integer spatial coordinates. Once the tri-planes generate a feature grid conditioned on the input image’s pose, a compact MLP decodes these features into a single-channel grayscale image, forming the corresponding 2D cross-section. RapidVol can also be initialized from a pre-computed fetal atlas. While this pre-initialization does not affect per-epoch reconstruction speed, it significantly accelerates convergence—achieving a structural similarity index (SSIM) of 0.9, 2.7$\times$ faster compared to random initialization. This method presents a promising avenue for speeding up 3D reconstruction and optimizing storage.

Research by \citet{wang_pixel2mesh_2018} introduces Pixel2MeshNet, a deep learning approach for creating 3D triangular mesh models from a single color image (Figure~\ref{fig:non-mri-reconstructions}b). Unlike prior methods which produced volume or point cloud representations, this research employs a graph deformation-based CNN to progressively refine an initial ellipsoid into a detailed 3D mesh. The network starts with an initial ellipsoid mesh with a smaller number of vertices then gradually increases the number of vertices and adds local details through graph unpooling layers. Each deformation block refines the mesh and extracts perceptual image features for the next block. This pipeline is particularly compelling for medical applications, as it demonstrates how mesh structure and surface topology can be learned and refined through a coarse-to-fine strategy using only 2D input. When applied to MRI reconstruction, similar graph-based mesh deformation techniques could be used to generate anatomically consistent surface meshes directly from 2D MRI slices or sparse 3D inputs, offering a lightweight and topologically-aware alternative to voxel-based methods, particularly in shape modeling tasks such as organ segmentation, cardiac surface generation, or cortical mapping.

An updated version of Pixel2MeshNet is Pixel2MeshNet++ \citep{wen_pixel2mesh_2019}, which iteratively refines a coarse initial shape by predicting a series of deformations. Pixel2MeshNet++ leverages cross-view information by sampling areas around mesh vertices and reasoning optimal deformations using perceptual feature statistics derived from multiple input images, mirroring traditional multi-view geometry techniques. While Pixel2MeshNet generates 3D mesh models from a single color image, Pixel2MeshNet++ is designed to generate object shapes from multiple color images with known camera poses. For each vertex of the current mesh, its multi-view deformation network samples a set of deformation hypotheses. For each hypothesized vertex location including the original vertex position, 3D vertex coordinates are projected onto 2D image planes using camera intrinsics and extrinsics to pool perceptual features from the input images. The authors highlight its ability to produce accurate, visually plausible 3D shapes that generalize well across various object categories, differing numbers of input images, and initial mesh quality. This coarse-to-fine, multi-view deformation strategy could be adapted for 3D MRI reconstruction by treating orthogonal or multi-slice MRI planes as different “views” of an anatomical structure. Leveraging such spatial context across views could enhance reconstruction accuracy, especially for complex, curved anatomical surfaces like the heart or cortex.

\section{Discussion}
In this review article, we presented a survey of recent literature on deep learning models, specifically CNNs, GANs, and diffusion models for researchers in medical imaging and deep learning. Our goal was to identify the most innovative and influential approaches that address the challenge of reconstructing 3D anatomical structures from 2D MRI slices. We carefully curated the list of available papers and models to select the most influential ones based on novelty, citations, and innovations. In particular, for each application of deep learning in MRI reconstruction we described the benefits and shortcomings. For each of these benefits, we provided a detailed and high-level abstraction of the core techniques. Moreover, we characterized the existing models based on dataset, organ, and model type (see Table~\ref{tab:research_papers}).

Compared to traditional approaches such as SSM, deep learning methodologies offer greater flexibility and scalability in capturing complex anatomical variability. While SSMs rely on linear combinations of pre-aligned templates and are limited by the quality and variability of training shapes, deep learning models can learn non-linear mappings from image data to 3D structures, enabling them to reconstruct shapes directly from raw MRI inputs. This allows for higher fidelity reconstructions, especially in cases with anatomical anomalies or incomplete data. Moreover, deep neural networks can integrate spatial context and semantic information through learned features, often outperforming traditional methods in both accuracy and robustness.

Our survey revealed that CNN-based models continue to dominate the field of 3D reconstruction, with architectures such as U-Net, 3D U-Net, and attention-based CNNs providing strong baselines for both segmentation and volumetric prediction. GCNs and recurrent networks have further expanded the capabilities of CNN-based systems by incorporating spatial and temporal dependencies. Recently, diffusion models like SADIR~\citep{jayakumar_sadir_2023} and DMCVR~\citep{he2023dmcvrmorphologyguideddiffusionmodel} have demonstrated promising performance improvements over earlier methods, offering better generalization and higher-quality reconstructions. GANs have gained traction due to their ability to produce realistic anatomical structures. Notably, models like nESRGAN~\citep{Zhangs21092978} use adversarial training to enhance 3D super-resolution outputs, significantly improving both perceptual quality and objective metrics. However, across all methods, the central concern remains ensuring anatomical plausibility and clinical utility.

As discussed, the choice of output representation is one of the most consequential decisions in 3D reconstruction model design. Different representations, such as volumetric occupancy grids, point clouds, meshes, and implicit functions, impose different constraints and enable different modeling strategies. Volumetric approaches, for example, provide dense internal detail and are compatible with traditional CNNs, but are computationally expensive. Surface-based methods like mesh reconstruction are more memory efficient and enable detailed surface modeling, but often depend on template deformations and suffer from topological constraints. Point cloud methods offer flexibility and simplicity but lack surface connectivity. Emerging hybrid models that combine multiple representations (e.g., point cloud-to-mesh networks) have shown promise in balancing expressiveness with tractability. Ultimately, the choice of representation should be guided by application-specific goals, such as segmentation accuracy, shape fidelity, or simulation compatibility.

A promising direction for future research lies in synthesizing the strengths of multiple model innovations across the literature. For example, the anatomical fidelity of shape-aware mesh deformation models~\citep{wen_pixel2mesh_2019, chen_shape_2021} could be enhanced by integrating the generative realism of diffusion-based networks~\citep{he2023dmcvrmorphologyguideddiffusionmodel, jayakumar_sadir_2023}. Similarly, the robustness of point cloud-to-mesh pipelines~\citep{beetz_point2mesh-net_2022} could be improved by embedding anatomical priors or latent diffusion guidance to better handle sparse clinical data. Hybrid architectures that combine volumetric prediction with surface refinement, or those that use transformer-based modules for long-range context aggregation, offer additional opportunities to build more expressive, anatomically consistent reconstructions. As the field matures, the convergence of these complementary strategies will be key to building generalizable, accurate, and clinically viable 3D MRI reconstruction systems.

A critical gap in current research is the lack of models trained on data from individuals with different pathologies. Despite their advantages, existing models often struggle to generalize beyond the training distribution. Most existing models are trained primarily on scans from healthy individuals, limiting their generalizability and reducing their clinical application. This bias poses a significant concern, as it excludes many patients who undergo MRI scans for diagnostic purposes. Addressing these data limitations requires a concerted effort to develop more robust, data-efficient models that can generalize across varied populations and disease states. There are an increasing number of available large-scale MRI datasets for different pathological conditions; a few are listed in Table~\ref{tab:public_mri_datasets}. Efforts to improve generalizability have included domain adaptation techniques, multi-site data harmonization, and data-efficient learning strategies. One promising direction is the use of shape-aware deformation models~\citep{wen_pixel2mesh_2019, chen_shape_2021, ye_neural_2024}, which integrate anatomical priors to guide reconstruction in cases where sparse or low-quality data is available. These models leverage geometric regularization and learned templates to adapt reconstructions to a wide range of anatomies, including those with disease-induced variability.

\begin{table}[!ht]
\centering
\caption{Publicly available MRI datasets featuring diseased pathologies.}
\sloppy
\begin{tabular}{|p{3.5cm}|p{2.5cm}|p{3.5cm}|p{3.5cm}|}
\hline
\textbf{Dataset Name} & \textbf{Country of Origin} & \textbf{Organ/Disease} & \textbf{Link} \\
\hline
Evaluation of Myocardial Infarction from Delayed-Enhancement Cardiac MRI (EMIDEC) & France & Heart / Myocardial Infarction & \url{https://emidec.com/dataset} \\
\hline
Automated Cardiac Diagnosis Challenge (ACDC) & France & Heart / Various Pathologies & \url{https://www.creatis.insa-lyon.fr/Challenge/acdc/index.html} \\
\hline
Multi-Disease, Multi-View \& Multi-Center
Right Ventricular Segmentation in Cardiac MRI (M\&Ms-2)
 & Spain & Heart / Various Pathologies & \url{https://www.ub.edu/mnms-2/} \\
\hline
Open Access Series of Imaging Studies (OASIS) & USA & Brain / Alzheimer's, Aging & \url{https://sites.wustl.edu/oasisbrains/} \\
\hline
Alzheimer's Disease Neuroimaging Initiative (ADNI) & USA & Brain / Alzheimer's & \url{https://adni.loni.usc.edu/data-samples/adni-data/} \\
\hline
Ischemic Stroke Lesion Segmentation (ISLES) Challenge & Multinational & Brain / Stroke Lesions & \url{https://www.isles-challenge.org/} \\
\hline
The Cancer Imaging Archive (TCIA) & USA & Multiple / Various Cancers & \url{https://www.cancerimagingarchive.net/} \\
\hline
\end{tabular}
\label{tab:public_mri_datasets}
\end{table}

Another challenging aspect is the extensive computational burden of state-of-the-art 3D reconstruction models. Diffusion models, for example, often require hundreds of sampling steps, while mesh deformation models can be sensitive to initialization and require careful tuning of loss weights. Research is needed to make these models more efficient, both in terms of training time and inference speed, especially for integration into real-time or resource-limited clinical environments. Promising strategies include multi-scale architectures and quantization \citep{li2023qdiffusionquantizingdiffusionmodels}.

\section{Conclusion}
This review article presents a comprehensive survey of deep learning techniques for 3D MRI reconstruction from 2D image stacks, emphasizing methods based on CNNs, GANs, and diffusion models. We explored a variety of output representations, including volumetric, surface-based, and point cloud formats, and analyzed their implications on model design, performance, and clinical utility.

Our analysis highlighted the trade-offs between different architectures and representations, the challenges of generalization to pathological cases, and the growing role of shape priors and generative modeling. As deep learning continues to evolve, future work should focus on integrating complementary innovations—such as combining mesh deformation with diffusion priors or fusing volumetric and surface-based predictions—to build more robust, interpretable, and clinically viable models. By providing an extensive view of this rapidly advancing field, this review will inform future research and help accelerate the deployment of 3D reconstruction methods in real-world clinical settings, providing a foundation for understanding the current landscape of deep learning in 3D MRI reconstruction.

\section{CRediT author statement}
\textbf{Emma McMillian:} Conceptualization, Methodology, Data Curation, Writing - Original Draft, Writing - Review \& Editing. \textbf{Abhirup Banerjee:} Supervision,  Data Curation, Writing - Review \& Editing. \textbf{Alfonso Bueno-Orovio:} Supervision, Data Curation, Writing - Review \& Editing.

\section{Funding Sources}
This work was supported by an Engineering and Physical Sciences Research Council Doctoral Award (CS2425{\textbackslash}1668004, to E.M.).
A.B. is supported by a Royal Society University Research Fellowship (URF{\textbackslash}R1{\textbackslash}221314). A.B.O. acknowledges support from the SMASH-HCM project (Innovate UK Grant 10110728).



\bibliographystyle{elsarticle-harv}
\bibliography{lit_review_references}


\newpage
\appendix
\section{Loss Functions}
This section explores common loss functions for training deep learning-based 3D reconstruction algorithms. The loss functions are categorized based on the model’s output representation. These categories include point, mesh-based, and volumetric losses for 3D reconstruction pipelines.

\subsection{Point-based Loss}
\subsubsection{Chamfer Distance}
For two point clouds \( P = \{p_i\}_{i=1}^{N} \) and \( Q = \{q_j\}_{j=1}^{M} \), the Chamfer Distance (CD) is
\begin{equation}
\mathcal{L}_{\text{CD}}(P, Q) = \frac{1}{|P|} \sum_{p \in P} \min_{q \in Q} \| p - q \|_2^2 + \frac{1}{|Q|} \sum_{q \in Q} \min_{p \in P} \| q - p \|_2^2.
\label{eq:chamfer}
\end{equation}

\subsubsection{Earth Mover’s Distance}
Given a bijection \( \phi: P \rightarrow Q \), the Earth Mover’s Distance (EMD) is
\begin{equation}
\mathcal{L}_{\text{EMD}}(P, Q) = \min_{\phi} \frac{1}{|P|} \sum_{p \in P} \| p - \phi(p) \|_2.
\label{eq:emd}
\end{equation}

\subsubsection{Point-wise L2 Loss}
When a one-to-one correspondence exists between points in \(P\) and \(Q\), the Point-wise L2 Loss is defined as
\begin{equation}
\mathcal{L}_{\text{L2}} = \frac{1}{N} \sum_{i=1}^{N} \| p_i - q_i \|_2^2.
\label{eq:pointwise_l2}
\end{equation}

\subsubsection{Hausdorff Distance}
Hausdorff Distance (HD) \citep{HDtran_chapter_2016} measures the largest minimum distance between two point sets, providing insight into local misalignments between predictions and ground truth. HD is defined mathematically as:
\begin{equation}
\mathcal{L}_{\text{HD}} = \max \left\{ \sup_{p \in P} \inf_{q \in Q} \| p - q \|_2, \sup_{q \in Q} \inf_{p \in P} \| q - p \|_2 \right\}.
\label{eq:hausdorff}
\end{equation}

\subsection{Mesh-based Loss}
\subsubsection{Chamfer Distance over Mesh Vertices}
For predicted and ground truth mesh vertex sets \( V_{\text{pred}} \) and \( V_{\text{gt}} \), the Chamfer Distance is defined as
\begin{equation}
\mathcal{L}_{\text{CD}} = \frac{1}{|V_{\text{pred}}|} \sum_{v \in V_{\text{pred}}} \min_{u \in V_{\text{gt}}} \| v - u \|_2^2 + \frac{1}{|V_{\text{gt}}|} \sum_{u \in V_{\text{gt}}} \min_{v \in V_{\text{pred}}} \| u - v \|_2^2.
\label{eq:mesh_cd}
\end{equation}

\subsubsection{Edge Length Regularization}
Given an edge set \( \mathcal{E} \) and vertex pairs \( (v_i, v_j) \in \mathcal{E} \), the Edge Length Regularization loss is defined as
\begin{equation}
\bar{l} = \frac{1}{|\mathcal{E}|} \sum_{(i, j) \in \mathcal{E}} \| v_i - v_j \|_2, \quad
\mathcal{L}_{\text{edge}} = \sum_{(i, j) \in \mathcal{E}} \left( \| v_i - v_j \|_2 - \bar{l} \right)^2.
\label{eq:edge_reg}
\end{equation}

\subsubsection{Laplacian Smoothness Loss}
For each vertex \( v_i \) with neighbors \( \mathcal{N}(i) \), the Laplacian Smoothness Loss is defined as
\begin{equation}
\mathcal{L}_{\text{lap}} = \sum_{i} \left\| v_i - \frac{1}{|\mathcal{N}(i)|} \sum_{j \in \mathcal{N}(i)} v_j \right\|_2^2.
\label{eq:laplacian}
\end{equation}

\subsubsection{Normal Loss}
For predicted and ground truth face normals \( n_i^{\text{pred}} \) and \( n_i^{\text{gt}} \), the Normal Loss is defined as
\begin{equation}
\mathcal{L}_{\text{normal}} = \sum_{i} \left(1 - \left\langle n_i^{\text{pred}}, n_i^{\text{gt}} \right\rangle \right).
\label{eq:normal}
\end{equation}

\subsection{Volumetric Loss}

\subsubsection{Binary Cross-Entropy Loss}
For predicted voxel probabilities \( \hat{y}_i \) and ground truth labels \( y_i \), the Binary Cross-Entropy (BCE) Loss is defined as
\begin{equation}
\mathcal{L}_{\text{BCE}} = - \frac{1}{N} \sum_{i=1}^{N} \left[ y_i \log(\hat{y}_i) + (1 - y_i) \log(1 - \hat{y}_i) \right].
\end{equation}

\subsubsection{Dice Loss / Dice Similarity Coefficient}
The Dice Loss, derived from the Dice Similarity Coefficient, balances false positives and false negatives and is particularly useful for imbalanced classes. It is defined as
\[
\text{Dice} = \frac{2 |\hat{Y} \cap Y|}{|\hat{Y}| + |Y|} = \frac{2 \sum_{i} \hat{y}_i y_i}{\sum_{i} \hat{y}_i + \sum_{i} y_i}.
\]
\[
\quad \Rightarrow \quad
\mathcal{L}_{\text{Dice}} = 1 - \text{Dice}.
\]

\subsubsection{Negative Intersection over Union}
The Negative Intersection over Union (IoU) Loss measures the overlap between predicted and ground truth voxels and is defined as
\[
\text{IoU} = \frac{\sum_i \hat{y}_i y_i}{\sum_i \hat{y}_i + \sum_i y_i - \sum_i \hat{y}_i y_i}.
\]
\[
\quad \Rightarrow \quad
\mathcal{L}_{\text{IoU}} = 1 - \text{IoU}.
\]

\subsubsection{Voxel-wise L1/L2}
The voxel-wise L1 Loss (Mean Absolute Error) calculated between ground truth and prediction is defined as
\begin{equation}
    \mathcal{L}_{\text{L1}} = \frac{1}{N} \sum_{i=1}^{N} | \hat{y}_i - y_i|
    \label{eq:voxelwisel1}.
\end{equation}

The voxel-wise L2 Loss (Mean Squared Error) calculated between ground truth and prediction is defined as
\begin{equation}
   \mathcal{L}_{\text{L2}} = \frac{1}{N} \sum_{i=1}^{N} (\hat{y}_i - y_i)^2.
    \label{eq:voxelwisel2}
\end{equation}

\section{Evaluation Metrics}
The most commonly used evaluation metrics for the task of 3D reconstruction are IoU for voxel representation and both CD and EMD for point cloud and mesh representations. Three other evaluation metrics are commonly used for 3D reconstruction quality assessment, particularly in voxel-based or segmentation-based methods: precision, recall, and F1 score. These metrics quantify the overlap between the predicted and ground truth volumes and are especially useful in evaluating models under class imbalance or in cases where small structures are difficult to reconstruct. \\

\subsection{Precision} 
Precision measures the proportion of correctly predicted positive voxels (true positives) out of all voxels predicted as positive. In 3D reconstruction, this quantifies how much of the predicted shape is actually accurate. High precision indicates few false positives, which is critical when over-segmentation must be avoided.
\begin{equation}
    \text{Precision} = \frac{TP}{TP + FP}.
    \label{eq:precision}
\end{equation}

\subsection{Recall} 
Recall measures the proportion of true positives captured by the model relative to all actual positives in the ground truth. In 3D reconstruction, recall reflects how completely the model captures the target anatomy. High recall is important when under-segmentation could omit essential structures.
\begin{equation}
    \text{Recall} = \frac{TP}{TP + FN}.
    \label{eq:recal}
\end{equation}

\subsection{F1 Score} 
The F1 score is the harmonic mean of precision and recall. It provides a single value that balances both over- and under-segmentation performance. It is particularly useful in 3D segmentation tasks with uneven class distribution, such as when reconstructing small or complex anatomical regions.
\begin{equation}
    \text{F1 Score} = \frac{2 \cdot \text{Precision} \cdot \text{Recall}}{\text{Precision} + \text{Recall}}.
    \label{eq:F1-score}
\end{equation}

Assessing the quality of 3D MRI reconstructions requires a combination of perceptual, structural, and geometric evaluation metrics to ensure accurate representations of anatomical features. Many studies use the same metrics but on different datasets, and vice versa, which complicates direct comparisons of performance. The metrics described below are general approaches that are commonly used to evaluate deep learning–based 3D MRI reconstruction.

One widely used metric is the structural similarity index measure (SSIM), which is a common perceptual similarity known for being more similar to human perception compared to metrics like peak signal-to-noise ratio (PSNR). PSNR quantifies voxel-wise intensity differences. Higher PSNR generally indicates lower reconstruction error. However, PSNR may not always reflect clinically relevant distortions. Often research combines both PSNR and SSIM \citep{U2024195} in quality assessment to be able to better align visual differences with numerical error, e.g. constructing 3D MRI reconstruction using MLP \citep{chen_computationally_2023}.

Beyond voxel-level comparisons, geometric accuracy is essential in 3D MRI evaluation for small anatomical shapes. Hausdorff distance (HD) measures the maximum discrepancy between corresponding points on reconstructed surfaces, providing insight into local misalignment between predictions and ground truth. HD is particularly useful for medical image applications where precise anatomical boundaries are critical in visualizations, e.g. reconstructing 3D meshes from point clouds \citep{chen_shape_2021}. Similarly, chamfer distance assesses the overall shape similarity between two point clouds, making it effective for evaluating reconstruction consistency in volumetric shapes. Dice similarity coefficient quantifies the overlap between reconstructions and ground truth, providing a robust measure of volumetric similarity, e.g. comparing planes and phases of cardiac shapes \citep{corral_acero_comprehensive_2024}. By incorporating multiple evaluation metrics, we can comprehensively assess the fidelity of 3D MRI reconstructions, balancing perceptual quality, structural accuracy, and geometric consistency.

\end{document}